\documentclass[10pt,twocolumn,letterpaper]{article}

\usepackage{iccv}              

\usepackage{pifont}
\usepackage{multirow}
\usepackage{siunitx}
\usepackage{colortbl}
\usepackage{xcolor}
\usepackage{bm}
\definecolor{first}{rgb}{1, 0.7, 0.7}
\definecolor{second}{rgb}{1,0.85, 0.7}
\definecolor{third}{rgb}{1,1, 0.8}

\newcommand{\parahead}[1]{\vspace{1mm}\noindent\textbf{{#1}.}\ }

\newcommand{\cmark}{\ding{51}}%
\newcommand{\xmark}{\ding{55}}%

\newcommand{\MethodName}{InfiniCube\xspace}

\newcommand{\Data}{{\mathbf{D}}}
\newcommand{\Latent}{{\mathbf{X}}}
\newcommand{\Cond}{{\mathbf{C}}}
\newcommand{\Real}{{\mathbb{R}}}
\newcommand{\Encoder}{{\mathcal{E}}}
\newcommand{\Decoder}{{\mathcal{D}}}
\newcommand{\Gauss}{{\mathcal{N}}}
\newcommand{\Identity}{\mathbf{I}}
\newcommand{\Mask}{{\mathbf{M}}}
\newcommand{\VoxelDepth}{{\mathbf{Z}}}
\newcommand{\FDA}{{\mathbf{F}_\text{DAV2}}}

\newcommand{\vx}{{\text{vx}}}
\newcommand{\video}{{\text{vd}}}

\newcommand{\rdagger}{{\color{red}$^\dagger$}}
\newcommand{\rddagger}{{\color{red}$^\ddagger$}}

\definecolor{myorange}{RGB}{253, 107, 99}
\definecolor{myred}{RGB}{192, 0, 0}

\definecolor{iccvblue}{rgb}{0.21,0.49,0.74}
\usepackage[pagebackref,breaklinks,colorlinks,allcolors=iccvblue]{hyperref}

\definecolor{color1}{HTML}{805d93} 
\definecolor{color2}{HTML}{ff715b} 
\definecolor{color3}{HTML}{f9cb40} 
\definecolor{color4}{HTML}{6cae75} 
\definecolor{color5}{HTML}{588bdd} 
\hypersetup{
    colorlinks=true,
    linkcolor=color2,      
    citecolor=color5,      
    filecolor=color2,      
    urlcolor=color2        
}

\usepackage{multicol}
\usepackage[capitalize]{cleveref}
\usepackage{wrapfig}
\crefname{section}{\S}{\S\S}
\crefname{subsection}{\S}{\S\S}
\crefname{conj}{Conj.}{Conj.}

\Crefname{assumption}{\textbf{H}\hspace{-3pt}}{\textbf{H}\hspace{-3pt}}
\crefname{assumption}{\textbf{H}}{\textbf{H}}

\crefname{algorithm}{\text{Alg.}}{\text{Alg.}}
\crefname{assumption}{\textbf{H}}{\textbf{H}}
\crefname{equation}{\text{Eq}}{\text{Eq}}
\crefname{definition}{\text{Dfn.}}{\text{Dfn.}}
\crefname{lemma}{\text{Lemma}}{\text{Lemma}}
\crefname{dfn}{\text{Dfn.}}{\text{Dfn.}}
\crefname{thm}{\text{Thm.}}{\text{Thm.}}
\crefname{tab}{\text{Tab.}}{\text{Tab.}}
\crefname{fig}{\text{Fig.}}{\text{Fig.}}
\crefname{table}{\text{Tab.}}{\text{Tab.}}
\crefname{figure}{\text{Fig.}}{\text{Fig.}}

\def\paperID{6529} 
\def\confName{ICCV}
\def\confYear{2025}

\title{\MethodName: Unbounded and Controllable Dynamic 3D Driving Scene Generation with World-Guided Video Models}

\author{
Yifan Lu$^{1,2}$\textcolor{red}{\footnotemark[1]}\quad Xuanchi Ren$^{1,3,4}$\textcolor{red}{\footnotemark[1]}\quad Jiawei Yang$^{5}$\quad Tianchang Shen$^{1,3,4}$\quad
Zhangjie Wu$^{1}$ \\ Jun Gao$^{1,3,4}$\quad Yue Wang$^{1,5}$\quad Siheng Chen$^{2}$\quad Mike Chen$^{1}$\quad Sanja Fidler$^{1,3,4}$\quad Jiahui Huang$^{1}$ \\ \vspace{-0.6em} \\
\small
$^{1}$NVIDIA, $^{2}$Shanghai Jiao Tong University, $^{3}$University of Toronto, $^{4}$Vector Institute, $^{5}$University of Southern California \\
\small\url{https://research.nvidia.com/labs/toronto-ai/infinicube/}\\
}

\begin{document}
\twocolumn[{%
\renewcommand\twocolumn[1][]{#1}%
\maketitle
\begin{center}
\renewcommand\arraystretch{0.5} 
\centering
\includegraphics[width=\linewidth]{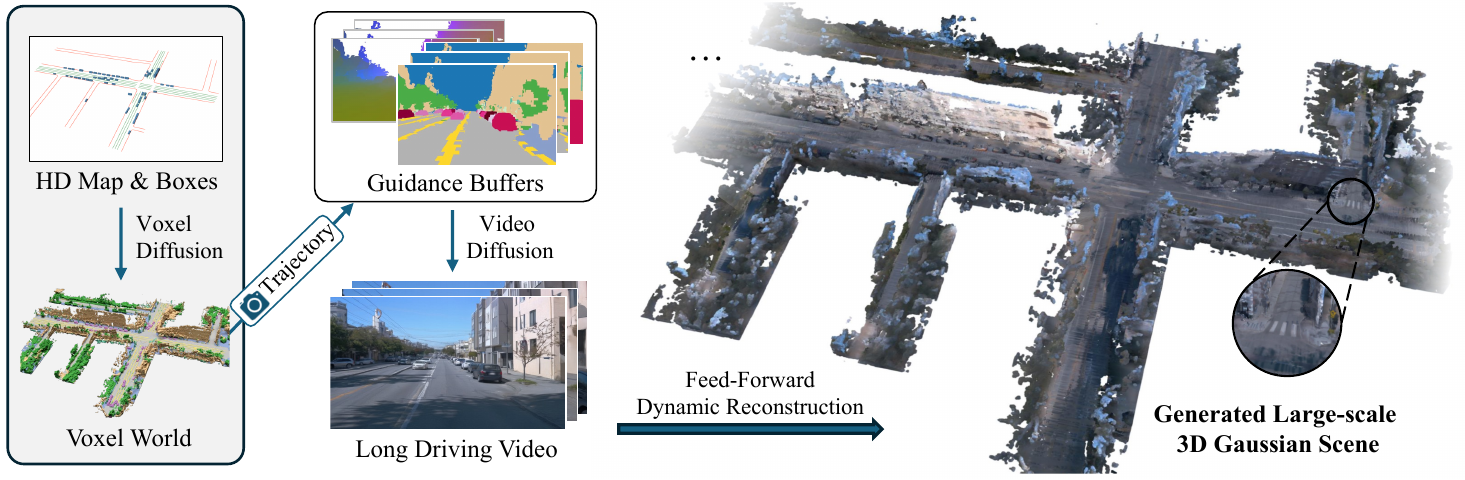}
\captionof{figure}{\textbf{\MethodName}. Our model generates large-scale dynamic 3D driving scenes given the HD map, 3D bounding boxes and text controls. Here we show a generated large-scale driving scene spanning $\sim$\SI{100000}{\meter\squared} in 3D Gaussians.}
\label{fig:teaser}
\end{center}
}]

\newcommand\blfootnote[1]{%
  \begingroup
  \renewcommand\thefootnote{}\footnote{#1}%
  \addtocounter{footnote}{-1}%
  \endgroup
}
\blfootnote{\textcolor{red}{*} Equal Contribution.}

\begin{abstract}
We present \MethodName, a scalable and controllable method to generate unbounded and dynamic 3D driving scenes with high fidelity.
Previous methods for scene generation are constrained either by their applicability to indoor scenes or by their lack of controllability.
In contrast, we take advantage of recent advances in 3D and video generative models to achieve large dynamic scene generation with flexible controls like HD maps, vehicle bounding boxes, and text descriptions.
First, we construct a map-conditioned 3D voxel generative model to unleash its power for unbounded voxel world generation. Then, we re-purpose a video model and ground it on the voxel world through a set of pixel-aligned guidance buffers, synthesizing a consistent appearance on long-video generation for large-scale scenes.
Finally, we propose a fast feed-forward approach that employs both voxel and pixel branches to lift videos to dynamic 3D Gaussians with controllable objects.
Our method generates realistic and dynamic 3D driving scenes, and extensive experiments validate the effectiveness of our model design. 

\end{abstract}    

\vspace{-1em}
\section{Introduction}
\label{sec:intro}

Generating simulatable and controllable 3D scenes is an essential task for a wide spectrum of applications, including mixed reality, robotics, and the training and testing of autonomous vehicles (AV)~\cite{hu2023planning,li2024hydra}.
In particular, the requirements of AV applications have introduced new challenges for 3D generative models in driving scenarios, posing the following key desiderata:
(1) \emph{fidelity and consistency}, to ensure that the generated scenes support photo-realistic rendering while preserving consistent appearance and geometry at large scales; and
(2) \emph{controllability}, to enables users to effortlessly specify scene layouts, appearances, and ego-car behaviors, facilitating the creation of diverse scenes tailored to their specific needs; and
(3) \emph{Representation in 3D}, to support reliable physics simulation like collision detection and LiDAR simulation.

The investigation into large-scale 3D driving scene generation that satisfies the above criteria has been an active research area. One type of approaches focuses on the direct learning of 3D priors. After encoding the 3D scene structure into either sparse voxels~\cite{lin2023infinicity,ren2024xcube} or neural fields~\cite{,kim2023neuralfield}, powerful generative models (\eg diffusion models~\cite{ho2020denoising}) are employed to model these encodings. While these approaches can generate valid 3D structures, they often fail to capture detailed appearance information and show low fidelity in rendering. 

Alternatively, the recent development of video generative models~\cite{blattmann2023stable,brooks2024video} has shown promising results in generating rich and high-fidelity visual details by pretraining on massive video datasets. Finetuning these models on driving datasets with additional conditions, such as High-Definition (HD) maps or bounding boxes, has demonstrated the capability of generating realistic videos~\cite{hu2023gaia,gao2024vista,gao2024magicdrivedit}.
However, there are two main problems with these methods: First, the generated videos often lack 3D consistency and are not represented in true 3D formats, which limits their direct applicability for 3D physical simulation tasks.
Second, existing video models generate a limited number of frames, making it difficult to provide long-distance and consistent footage for large-scale scenes.



In this paper, we identify the key challenges in the above two types of methods, and 
present \textbf{\MethodName} (\cref{fig:teaser}), a novel scalable method that generates \emph{large-scale dynamic 3D Gaussian scenes} with high fidelity.
Specifically, we first construct a map-conditioned 3D voxel generative model, and generate unbounded voxel worlds through outpainting.
The generated voxel world is subsequently rendered into a set of guidance buffers to assist in long video generation for appearance synthesis with the video model.
We further propose a fast and robust method to lift the video and the voxels to dynamic 3D Gaussians (3DGS)~\cite{kerbl20233d} while preserving the controllability of dynamic vehicles. In \cref{table:method_compare}, we provide a schematic comparison of our method with other solutions: \MethodName not only enjoys long video generation but also scales to large 3D dynamic scenes up to $\sim$\SI{100000}{\meter\squared} (around \qtyproduct{300x400}{\meter}), providing support for various applications based upon the real 3D representation. We summarize our contribution as follows.
\begin{itemize}
    \item We introduce a new pipeline for high-fidelity, and large-scale dynamic 3DGS scene generation conditioned on HD maps, vehicle bounding boxes and text prompts.
    \item We realize unbounded voxel world generation with outpainting and ensure high consistency between chunks.
    \item We innovatively propose using guidance buffers rendered from voxel worlds to extend video lengths for large-scale scenes, increasing the number of high-quality frames to 200 from original 25-frame Stable Video Diffusion XT.
    \item We present a novel dual-branch feed-forward reconstruction method capable of generating dynamic 3DGS scenes from dynamic videos within just few seconds.
\end{itemize}

\begin{table}[!t]
\footnotesize
\setlength{\tabcolsep}{2.5pt}
\begin{center}
\begin{tabular}{lcccc}
\toprule
\multirow{2}{*}{} & \multicolumn{2}{c}{Output Type} & \multirow{2}{*}[-0.3em]{\begin{tabular}[c]{c}Detailed\\ Geometry\end{tabular}} & \multirow{2}{*}[-0.3em]{\begin{tabular}[c]{c}Driving\\ Length\rdagger \end{tabular}} \\ \cmidrule(lr){2-3}
& \hspace{0.5em} Video    & {3D Rep.}  &    &    \\ \midrule
Vista~\cite{gao2024vista} & \hspace{0.5em} \cmark & \xmark  & \xmark & \qty{15}{\second}                 \\
MagicDrive3D~\cite{gao2024magicdrive3d}  & \hspace{0.5em} \cmark   & 3DGS  & \cmark  & \qty{6}{\second} \\
InfiniCity~\cite{lin2023infinicity}    & \hspace{0.5em} \xmark  & Voxels    & \xmark   & \texttt{N/A}   \\
WoVoGen~\cite{lu2025wovogen}  & \hspace{0.5em} \cmark  & Voxels    & \cmark  & \qty{0.4}{\second}  \\ \midrule
\MethodName (\textbf{Ours})  & \hspace{0.5em} \cmark  & Voxels + 3DGS    & \cmark & \qty{20}{\second}\rddagger \\ \bottomrule
\end{tabular}
\end{center}
\vspace{-1em}
\rdagger: Measuring the video model generation length at 10Hz. \\
\rddagger: Our method allows unlimited free driving in our 3DGS scene.
\vspace{-1em}

\caption{\textbf{High-level comparison with existing solutions.} Our method outputs both video and a renderable 3D representation (`3D Rep.') with detailed geometry and a longer driving length.}
\label{table:method_compare}
\vspace{-2em}
\end{table}
\vspace{-1em}

\section{Related Work}
\vspace{-1em}
\label{sec:rw}
\parahead{3D Generation}
Training 3D generative models for object-level shapes has witnessed significant progress in recent years, with either directly learning the 3D shape distribution~\cite{xie2024latte3d,gao2022get3d,zhang2024clay}, or first generating multi-view images and then lifting them to 3D~\cite{shi2023mvdream,xu2024grm,gao2024cat3d}.
Extending these techniques to larger-scale scenes is however non-trivial due to its high complexity.
Some works directly learn the 3D scene distribution~\cite{ren2024xcube,zyrianov2024lidardm,kim2023neuralfield,liu2023pyramid,zhang2024urban} with carefully curated ground-truth data, while some others take the combinatorial~\cite{li2024dreamscene,gao2024graphdreamer} or hierarchical~\cite{lin2023infinicity,xie2024citydreamer} approach of generating or retrieving objects and then arranging them using guided scene layouts.
Notably, another emerging trend is to generate 3D scenes by reconstruction from video model's outputs~\cite{yu20244real,zhang20244diffusion} thanks to their rich appearance.

\parahead{Controllable Video Generation}
The prevalence of video generation models fueled by either diffusion models~\cite{blattmann2023align,blattmann2023stable,chen2024videocrafter2} or autoregressive models~\cite{yan2021videogpt,kondratyuk2023videopoet,RenW22} has brought up the need for more fine-grained control over the generated content, including camera trajectories~\cite{kuang2024collaborative,bahmani2024vd3d}, object motions~\cite{wang2024motionctrl,shi2024motion,yin2023dragnuwa}, and scene structures~\cite{liu2024reconx,guo2025sparsectrl}.
In the more specific domain of driving video generation, several works have conditioned the video generation on the HD maps and the car bounding boxes~\cite{yang2024drivearena,wen2024panacea,wang2024driving,hu2023gaia,gao2024vista, gao2024magicdrivedit}, and can generate a local occupancy map alongside~\cite{lu2025wovogen,gu2024dome}, turning them into a ``world model'' for planning.  However, existing video models suffer from a limited frame number for large-scale scenes, while InfiniCube achieves a much longer video length with 3D grounded guidance buffers.

\parahead{Driving Scene Reconstruction}%
Driving scene reconstruction plays a critical role in creating realistic simulation environment for autonomous vehicle. 
Existing methods, such as those based on neural radiance fields (NeRFs)~\cite{guo2023streetsurf,wu2023mars,yang2023emernerf,tonderski2024neurad} or 3DGS~\cite{chen2024omnire,fischer2024dynamic,yan2024street,zhou2024drivinggaussian}, achieve impressive visual fidelity but suffer from long training times, limiting their application at scale.
A concurrent trend is the development of large reconstruction models that leverage data priors for improved generalizability and fast inference~\cite{ren2024scube,yang2024storm,zhang2025gs}, which allows the reconstruction of 3D scenes in a few seconds. InfiniCube also adopts such a feed-forward strategy, tailored to account for the presence of dynamic objects with a novel dual-branch reconstruction method.

\section{Preliminaries}
\label{sec:prelim}

Our method is based on the following key concepts:

\parahead{Latent Diffusion Models (LDM)}%
Latent diffusion models~\cite{rombach2022high} are a class of generative models that learn the distribution from the latent $\Latent$ of the data $\Data$.
It is used together with an auto-encoder (or a tokenizer) that maps the data into the latent space $\Latent = \Encoder (\Data)$.
A diffusion model~\cite{ho2020denoising} starts with a noise vector $\Gauss(0, \Identity)$ and iteratively denoises it to generate a sample $\Latent$, potentially guided by a condition $\Cond$.
The decoder is eventually applied to the denoised latent to generate the output data $\hat{\Data} = \Decoder (\Latent)$.
LDMs have been shown to be effective and efficient in modeling many data modalities, including images~\cite{rombach2022high}, videos~\cite{blattmann2023stable}, and 3D~\cite{zhang2024clay}.

\parahead{Sparse Voxel LDM}%
As the 3D equivalent of pixels, voxels are uniform and structured and are hence suitable in deep learning applications.
In practice, only voxels intersecting actual surfaces need to be stored, forming a \emph{sparse} voxel grid.
In \MethodName, we consider the sparse voxel grid that stores a \emph{semantic label} in each of its voxels, represented as $\Data^\vx$.
The work of XCube~\cite{ren2024xcube} that we base on provides an efficient way of encoding sparse voxels $\Data^\vx$ into a dense latent feature cube $\Latent^\vx = \Encoder^\vx (\Data^\vx) \in \Real^{N^3 \times C}$ (where $N$ is the edge length), and decoding it back with high fidelity. In InfiniCube, we further condition on maps and vehicle boxes and realize unbounded voxel scene generation.
\section{Method}
\label{sec:method}
\begin{figure*}
    \vspace{-1em}
    \centering
    \includegraphics[width=\linewidth]{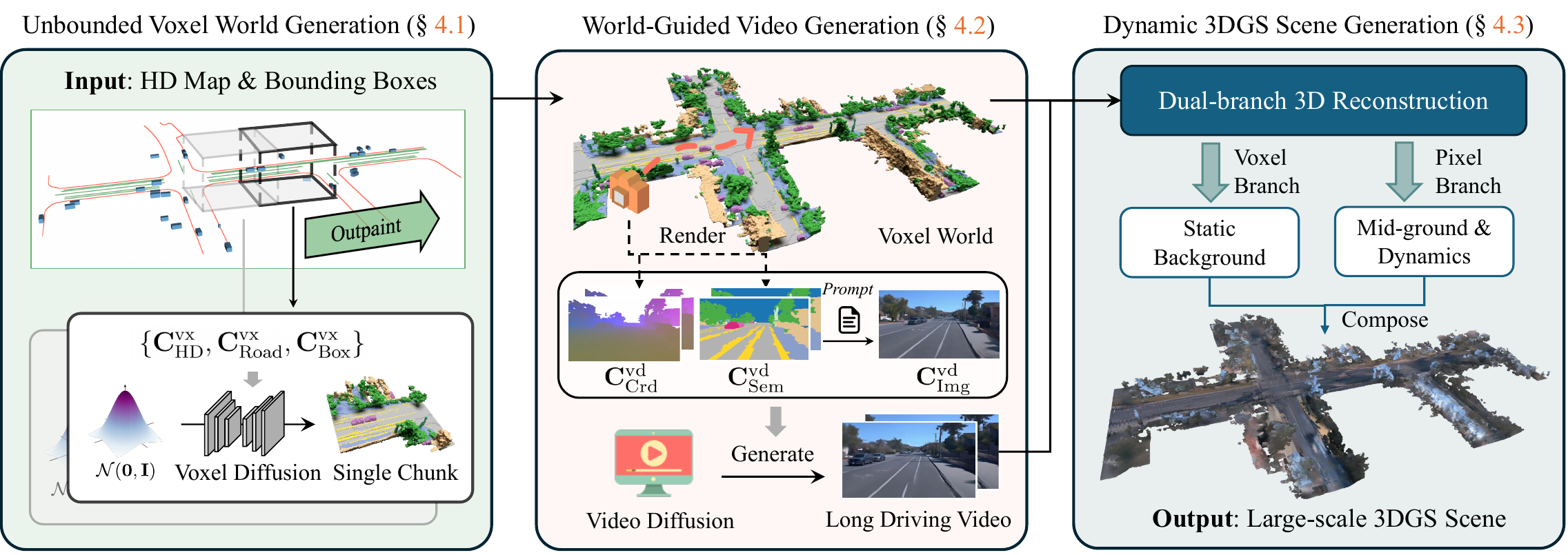}
    \vspace{-1.5em}
    \caption{\textbf{Pipeline.} Conditioned on HD maps and bounding boxes, we first generate a 3D voxel world representation. We then render the voxel world into several guidance buffers to boost video generation. The generated video and voxel world are jointly fed into a feed-forward dynamic reconstruction module to obtain the final 3DGS representation.}
    \label{fig:pipeline}
    \vspace{-1.5em}
\end{figure*}

\MethodName aims to generate large-scale dynamic 3D scenes guided by the input HD maps, vehicle bounding boxes and text prompts.
As shown in \cref{fig:pipeline}, we first generate a large-scale semantic voxel world of the target scene (\cref{subsec:method:voxel}) based on map and box conditions.
Such a world representation is then used to render several \emph{guidance buffers} on the given vehicle trajectories to support long-range video generation with textual controls(\cref{subsec:method:video}).
Finally, we take both the voxels and the synthesized videos to reconstruct a dynamic scene with the 3DGS representation (\cref{subsec:method:reconstruction}).

\subsection{Unbounded Voxel World Generation}
\label{subsec:method:voxel}

This step takes the HD map and the vehicle bounding boxes as input and synthesizes a corresponding 3D voxel world with semantic labels.
We develop the sparse voxel LDM based on XCube~\cite{ren2024xcube} (\cref{sec:prelim}) for this task, but extend it to unbound voxel scene generation by outpainting.

\begin{figure}
\centering
\includegraphics[width=\linewidth]{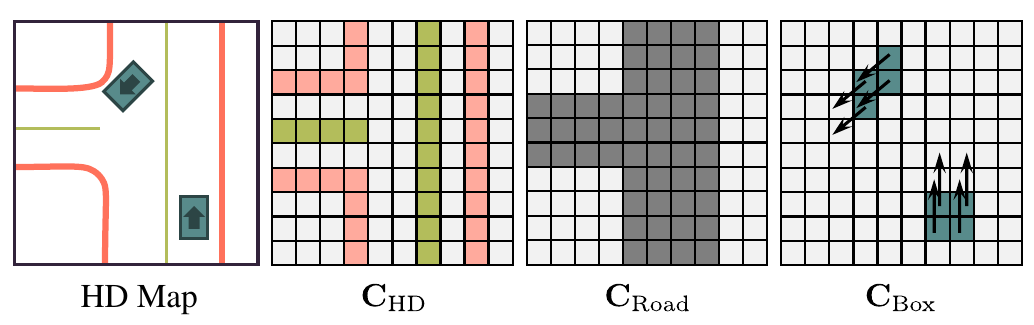}
\vspace{-1.5em}
\caption{\textbf{Conditions for Voxel World Generation.} 
We illustrate the conditions in 2D for clarity but the actual conditions are in 3D.}
\vspace{-1.5em}
\label{fig:cond-3d}
\end{figure}

\parahead{Building Map Conditions}
From our input, we first build a condition volume $\Cond^\vx \in \Real^{N^3 \times S}$ that shares the same structure as $\Latent^\vx$.
It contains the following three components.
\emph{(i)} \textbf{HD Map Condition} $\Cond_\text{HD}^\vx$:
Our input HD Map contains two sets of 3D polylines, road \emph{edges} defining the road boundary and road \emph{lines} that separate the lanes.
We rasterize the polylines into two separate channels of the condition $\Cond_\text{HD} \in \Real^{N^3 \times 2}$, with the voxel value set to 1 if any part of the voxel intersects with the polyline, and 0 otherwise.
\emph{(ii)} \textbf{Road Surface Condition} $\Cond_\text{Road}^\vx$:
We empirically find it difficult for the model to determine the drivable road regions from $\Cond_\text{HD}^\vx$ when the HD map is partly included in the condition $\Cond_\text{HD}$, which can happen in a local chunk generation.
We hence add another condition $\Cond_\text{Road}^\vx \in \Real^{N^3 \times 1}$ that delineates the voxelized road \emph{surface} as an extra signal to inform the model of the drivable area. We achieve this by fitting 3D planes with given road edges and road lines, and keeping their road surface parts.
\emph{(iii)} \textbf{Bounding Box Condition} $\Cond_\text{Box}^\vx$:
Bounding boxes contain detailed information about the vehicles' poses.
However, naïvely voxelizing bounding box occupancies can lead to information lost due to the coarse latent voxel size (\eg \SI{1.6}{\meter}).
We hence encode the vehicles' heading angles $\alpha$ as two-channel vectors $\left[ \sin\alpha, \cos\alpha \right]$, and set the feature in $\Cond_\text{Box}^\vx \in \Real^{N^3 \times 2}$ to this encoding if more than half of the corresponding voxel is occupied by the bounding box.
The full condition volume is a concatenation of the three $\Cond^\vx = \{ \Cond_\text{HD}^\vx, \Cond_\text{Road}^\vx, \Cond_\text{Box}^\vx \}$ with $S=4$, as visualized in \cref{fig:cond-3d}.

\parahead{Single Chunk Generation}
With the above $\Cond^\vx$, we apply the diffusion sampling procedure in \cite{ren2024xcube} to generate our desired semantic voxel grid representation $\Data^\vx$.
In a nutshell, the sampling process starts with a Gaussian white noise $\Gauss(0, \Identity)$ in shape $\Real^{N^3 \times C}$ which is subsequently concatenated with the conditions $\Cond^\vx$ in the channel dimension.
A 3D U-Net will then iteratively denoise the white noise to obtain the latent $\Latent^\vx$, which could be decoded to the desired semantic sparse voxel grid. We do not apply hierarchical upsampling like XCube~\cite{ren2024xcube} here. More details about the sampling procedure can be found in the Supplement.

However, a single pass of the LDM sampling can only generate one chunk of the scene. We hence propose the following strategy to \emph{outpaint} the grid to a large voxel world.

\parahead{Unbounded Scene Outpainting}
Here we use a strategy similar to Repaint~\cite{lugmayr2022repaint}, a training-free outpainting technique that is commonly used in diffusion models, to iteratively extend the scene in a seamless manner.
Specifically, during the generation of a new chunk, we ensure its sufficient overlap with the existing part of the scene, and take the latent $\Latent_\text{exist}^\vx$ from the overlapping area.
During the diffusion process, we keep $\Latent_\text{exist}^\vx$ fixed and only update the current latent $\Latent_\text{new}^\vx$ for the newly generated part, as follows:
\begin{equation}
    \Latent_\text{new}^\vx \leftarrow (1 - \Mask) \odot \hat{\Latent}_\text{new}^\vx + \Mask \odot \hat{\Latent}_\text{exist}^\vx,
\end{equation}
where $\Mask$ is the overlapping mask and $\odot$ is the element-wise product. 
While $\hat{\Latent}_\text{new}$ is sampled from the learned diffusion posterior, $\hat{\Latent}_\text{exist}$ is sampled from the noised version of the fixed $\Latent_\text{exist}$. With this, we can generate a large-scale scene that is consistent across different chunks.

\subsection{World-Guided Video Generation} 
\label{subsec:method:video}

We choose Stable Video Diffusion XT (SVD-XT)~\cite{blattmann2023stable} as our base model for video generation.
Under the hood, the video $\Data^\video$ is represented as a volume of shape $\Real^{H\times W \times T \times 3}$, where $H, W, T$ are the height, width, and number of frames, respectively.
The latent of the video modeled by the LDM is $\Latent^\video = \Encoder^\video (\Data^\video) \in \Real^{h \times w \times T \times 4}$, downsampling the spatial resolution with $h=\frac{H}{8}$ and $w=\frac{W}{8}$.
Similarly to the voxel generation model, the condition to the model $\Cond^\video \in \Real^{h \times w \times T \times M}$ shares the same size as $\Latent^\video$ except for the last channel dimension $M$.
The official model of SVD is trained on large-scale Internet videos, and supports conditioning on the first video frame with $\Cond_\text{Img}^\video  \in \Real^{h \times w \times T \times 4}$, derived by repeating the frame $T$ times and then encoding it with the SVD encoder $\Encoder^\video(\cdot)$.
To generate long videos over $T$ frames, we reuse the latent of the last generated frame as condition and rerun the inference pass in an auto-regressive manner, which is a training-free strategy for long-video generation.

However, generating long and consistent driving videos in this way is challenging since the accumulative error will significantly decline the frame quality, especially when performing auto-regression over 3 passes. To tackle this challenge, we propose to use \textbf{3D grounded guidance buffers} rendered from voxel worlds (from \cref{subsec:method:voxel}) to assist the video model in better capturing vehicle motion and environmental changes. They offer vital 3D grounding and guidance for the video model, significantly alleviating the accumulative error when generating a long video sequence.

\parahead{Guidance Buffers}%
Our guidance buffers are renderings of the 3D world into the video frames with given camera parameters.
They are composed of the following components.
\emph{(i)} The \textbf{Semantic Buffer} $\Cond^\video_\text{Sem}$ is the rendering of the voxel world's semantic labels.
We design a fixed discrete color palette to map the semantic labels to RGB values (see Supplement).
The color palette is chosen to add distinction between different semantic categories and fits the value range of the pretraiend encoder $\Encoder^\video(\cdot)$.
Notably, to differentiate between vehicle instances that belong to the same semantic category, we assign different saturation levels to their rendered colors. This enables the video model to maintain the distinct appearance of individual cars when there are multiple cars overlapping each other in the buffer.
\emph{(ii)} The \textbf{Coordinate Buffer} $\Cond^\video_\text{Crd}$ contains the 3D coordinates of the first voxel that each pixel ray hits (\cf \cite{wang2024dust3r,wang2019normalized}). 
For the same locations in the 3D scene across different frames, although they may be projected to different pixel coordinates due to ego and object motion, their pixel values (3D voxel coordinates) remain consistent.
For a $T$-frame video model, we further gather all the hit 3D voxels from $T$ frames, normalize their coordinates by a constant, and clamp the range to $[-1, 1]$, which will be accepted by $\Encoder^\video(\cdot)$. 
Coordinate buffer helps establish scene correspondences across frames and is useful when the semantic buffer exhibits a repeating pattern. 
A visualization of the guidance buffers can be found in \cref{fig:pipeline}.
Together the channel size of the video model condition $\Cond^\video = \{ \Cond^\video_\text{Img}, \Cond^\video_\text{Sem}, \Cond^\video_\text{Crd} \}$ is $M=12$.

Our guidance buffers can be built very efficiently by rendering the voxel world using $f$VDB~\cite{williams2024fvdb}.
Compared to other conditioning strategies such as pose embedding~\cite{gao2024magicdrive3d} or displacements~\cite{gao2024vista}, our method is agnostic to the metric scale of the trajectory and can achieve precise controls.

\parahead{Adding Text Prompts}%
Our video model necessitates a image as the condition. To achieve this, we train a ControlNet~\cite{zhang2023adding} based on FLUX~\cite{flux} to generate the initial frame using semantic buffers as control images. This approach allows us to incorporate text descriptions for scene generation while enhancing the quality and diversity of the video content by leveraging a pre-trained image diffusion model.

\subsection{Dynamic 3DGS Scene Generation}
\label{subsec:method:reconstruction}
While the above world-guided video model can already provide a photo-realistic and consistent appearance, a 3D representation is still important and required by simulation tasks. Therefore, we present a novel feed-forward method that is deeply integrated into our pipeline to reconstruct a 3DGS~\cite{kerbl20233d} scene with dynamic objects.

\begin{figure}
    \centering
    \includegraphics[width=\linewidth]{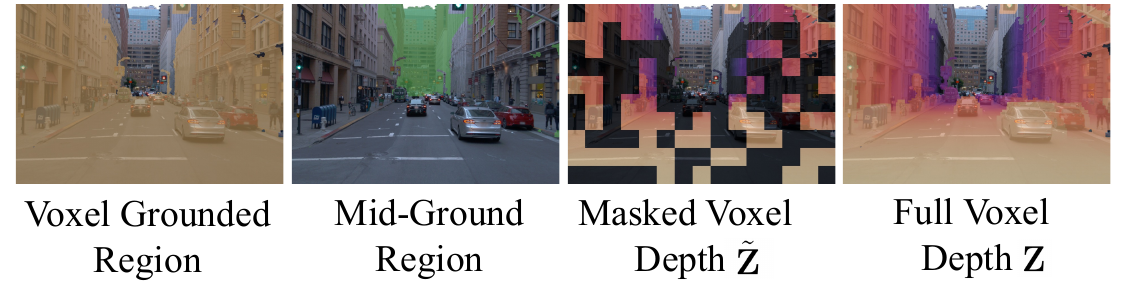}
    \vspace{-1.2em}
    \caption{\textbf{Illustration of concepts in the pixel branch.} The mid-ground region and masked / full voxel depth.}
    \label{fig:illustrate_gsm}
    \vspace{-1.2em}
\end{figure}

State-of-the-art feed-forward scene reconstruction models that leverage the 3DGS representation infer Gaussian attributes either in the \emph{voxel space}~\cite{ren2024scube} or in the \emph{pixel space}~\cite{zhang2025gs}.
While the former typically has a better geometry distribution given voxel scaffolds, the latter could better capture contents in the \emph{mid-ground} areas and the per-frame movement for the dynamic objects.
Here we elaborate \emph{mid-ground} as the pixel regions that (1) have no overlap with the projected voxels, and (2) do not belong to the sky (visualized in \cref{fig:illustrate_gsm}).
We hence propose a new \emph{dual-branch} reconstruction method that combines both advantages for our large-scale dynamic 3DGS scene generation. 

\parahead{Voxel Branch}%
One branch of our model takes the voxel world and the posed video frames as input, and outputs a set of 3D Gaussians for each voxel.
Similar to the appearance reconstruction branch in SCube~\cite{ren2024scube}, we unproject the features of the images to the voxel world and then apply a 3D sparse convolution U-Net architecture to transform the features into the per-voxel Gaussian attributes. Note that we mask the dynamic objects out from the image features and only use the static background voxels in this branch. 

\parahead{Pixel Branch}%
Our pixel branch employs a 2D UNet~\cite{unet} backbone to convert input images into per-pixel 3D Gaussians, similar to GS-LRM~\cite{zhang2025gs}, which simplifies the 3D-lifting task to a depth estimation problem. 
Here, we further propose \textbf{a self-supervised training strategy} to enhance the depth prediction capability and generalizability using rendered voxel depth $\VoxelDepth$ in this branch.
Specifically, we incorporate $\VoxelDepth$ as both an input and a supervisory signal:
During training, we use a randomly masked version of $\VoxelDepth$ (denoted as $\tilde{\VoxelDepth}$) to simulate the region that is not grounded by voxels, and supervise the predicted depth with the full $\VoxelDepth$ --- see the illustration in \cref{fig:illustrate_gsm}. This helps our pixel branch predict reasonable mid-ground depth at inference time.
We also supplement the network with the ViT backbone features $\FDA$ from a state-of-the-art depth estimation model Depth Anything V2~\cite{yang2024depth}.
In summary, the network takes the input images, $\tilde{\VoxelDepth}$, and $\FDA$ as input, and outputs the per-pixel 3DGS attributes (color, rotation, depth, scale, \etc).

\parahead{Sky Modeling}%
Modeling the sky region at infinite depth is challenging since most of it is not visible from the images.
We hence adopt an implicit sky representation from STORM~\cite{yang2024storm} that is highly generalizable to unseen regions.
Specifically, we use a light-weight encoder to summarize a single sky feature vector $\mathbf{c} \in \Real^{192}$ from the images, and use AdaLN~\cite{karras2019style} to modulate a Multi-Layer Perceptron (MLP) that takes a viewing angle and outputs RGB colors. 
More network details can be found in the Supplement.

\parahead{Supervision}%
We train two branches separately using photometric loss.
For the pixel branch, we additionally add the aforementioned depth loss. 

\parahead{Inference with Dynamic Objects}%
During inference, the voxel branch is applied only to the static part of the scene.
The pixel branch is applied iteratively for every 4 frames, but we only keep the 3DGS corresponding to the pixels of mid-ground regions and dynamic objects. 
Notably, for the dynamic vehicles, we extract the 3DGS belonging to each \emph{individual object} using the segmentation from the {Semantic Buffer} in \cref{subsec:method:video}, transform and aggregate them using their poses, and keep Gaussians in input bounding boxes to composite their final Gaussians. 
The motions of the objects are fully controllable by simply altering their trajectories.

\begin{figure*}[h]
    \vspace{-1.3em}
    \centering
    \includegraphics[width=\linewidth]{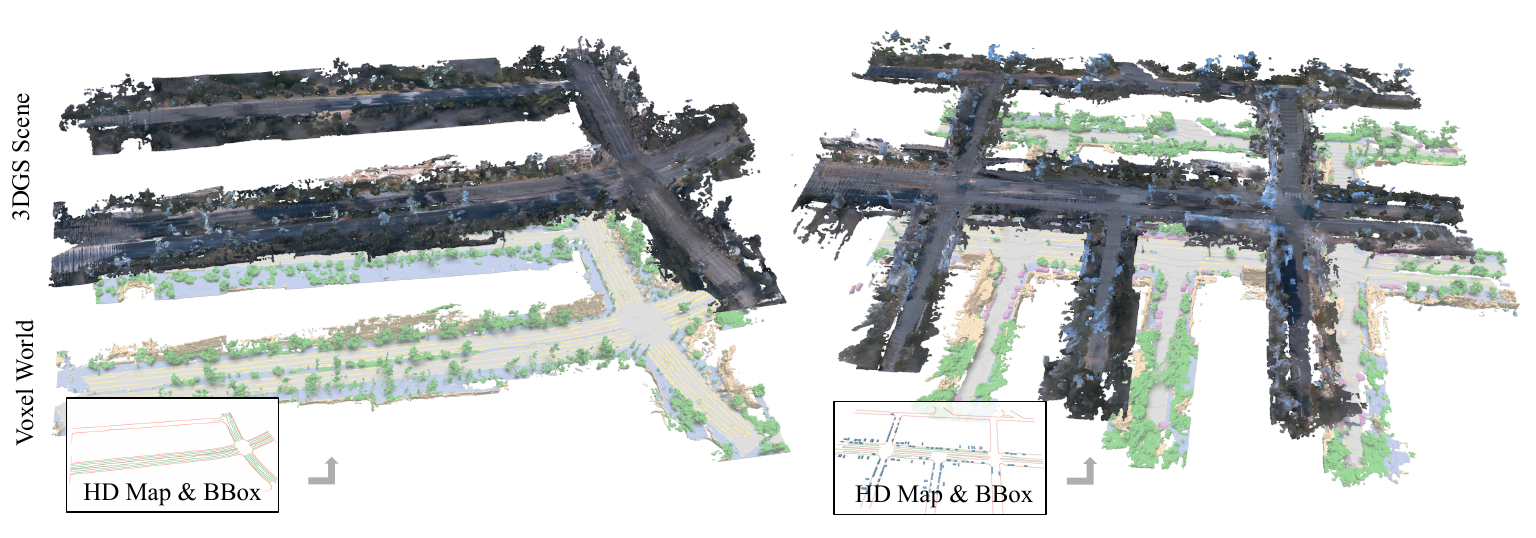}
    \vspace{-1.7em}
    \caption{\textbf{Full pipeline results}. The generated 3DGS scene and voxel world adhere to the input HD map and bounding boxes and extend over hundreds of meters with fine details. The text prompt is \textit{"Daytime, Sunny with a bright blue sky"}.}
    \label{fig:hd_map}
    \vspace{-1.2em}
\end{figure*}

\vspace{-0.5em}
\section{Experiments}
\label{sec:exp}

\subsection{Data Processing}%
Our model is trained on Waymo Open Dataset~\cite{sun2020scalability}, which provides LiDAR data, images, and accurate annotations of HD maps and vehicle bounding boxes.
To extract the ground-truth scene geometry to supervise the semantic voxel generation, we follow the approach of \cite{ren2024scube} by combining accumulated LiDAR points and the dense geometry obtained from the multi-view stereo pipeline of COLMAP~\cite{schonberger2016structure}.
The geometry of the dynamic cars is accumulated from the LiDAR points in their canonical space defined by their bounding box trajectories.
To enable text prompt conditioning, we annotate the video frames with textual descriptions. We employ Llama-3.2-90B-Vision-Instruct~\cite{dubey2024llama} to summarize the weather and time of day based on a stitched thumbnail created from 8 timestamps across 3 different views.

For each data sample used in training the voxel generation stage, we crop and voxelize the extracted geometry into a local chunk of 
\qtyproduct{51.2 x 51.2}{\meter} with a voxel size of \qty{0.2}{\meter}, centered around a randomly sampled ego-vehicle pose. 
We remove low-quality voxel grids and sequences with mostly static ego trajectories, resulting in 618 sequences for training and 90 sequences for evaluation.

\subsection{Implementation Details}

For both the voxel world (\cref{subsec:method:voxel}) and the 3DGS scene (\cref{subsec:method:reconstruction}) generation stages where a 3D sparse network backbone is needed, we take heavy use of the $f$VDB~\cite{williams2024fvdb} framework, and design our 3D auto-encoder and diffusion backbone similar to \cite{ren2024xcube,ren2024scube}. Please refer to the Supplement for network architecture details. For the video generation stage (\cref{subsec:method:video}), we use 25-frame SVD-XT~\cite{blattmann2023stable} base model from \texttt{\href{https://huggingface.co/stabilityai/stable-video-diffusion-img2vid-xt}{diffusers}}~\cite{diffusers} library. 
We finetune the pretrained model with a resolution of $576 \times 1024$ together with our designed conditions. 
We add Gaussian noise augmentation to perturb the conditioning latent features to improve the model robustness. 
During inference, we set the classifier-free guidance~\cite{CFG} weight to $3.0$ and use a denoising step of $25$. 
The voxel generation stage is trained for 48 GPU days, the video generation stage is trained for 192 GPU days, and both the voxel and pixel branches of the scene generation stage are trained for 32 GPU days, all using NVIDIA A100 GPUs.

\subsection{Large-scale Dynamic Scene Generation}
\begin{figure}
    \centering
    \includegraphics[width=\linewidth]{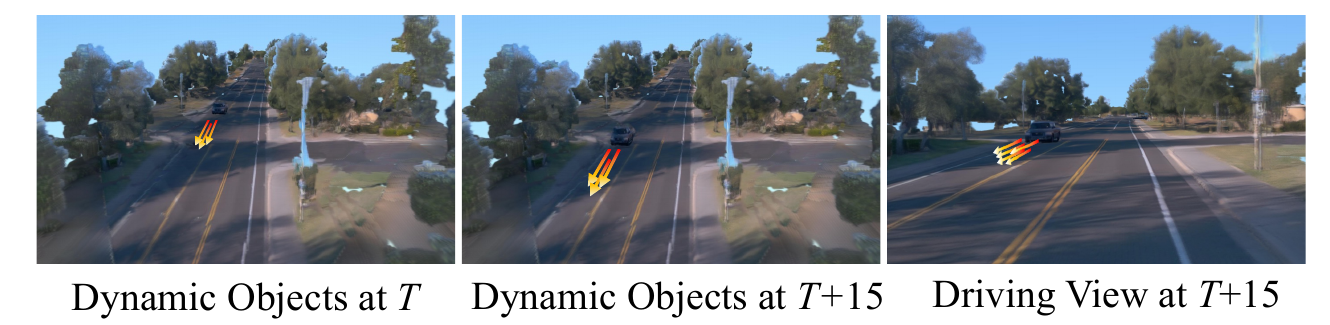}
    \vspace{-1.8em}
    \caption{\textbf{Dynamic 3DGS visualization.} We generate dynamic 3DGS with full controllability of dynamic objects across frames.}
    \label{fig:dynamic_recon}
    \vspace{-0.5em}
\end{figure}

We visualize the generated scenes from our full pipeline in \cref{fig:teaser} and \cref{fig:hd_map} .
Furthermore, \cref{fig:dynamic_recon} shows a close-up view of the dynamic objects, where a dynamic vehicle moves from frame $T$ to $T+15$.
Given just the HD map with 3D bounding boxes, our method can generate a complete scene with a high-fidelity appearance and controllable actors.
The scene is also rich in details and accurate in geometry, allowing for large-scale bird-eye visualizations that no prior work could generate.
These results are made possible by the synergy among the proposed components (\cref{subsec:method:voxel,subsec:method:video,subsec:method:reconstruction}). In the following sections, we will analyze their importance in detail.

\subsection{Main Components Analysis}

\subsubsection{Voxel World Generation}

\begin{figure}
    \centering
    \includegraphics[width=0.95\linewidth]{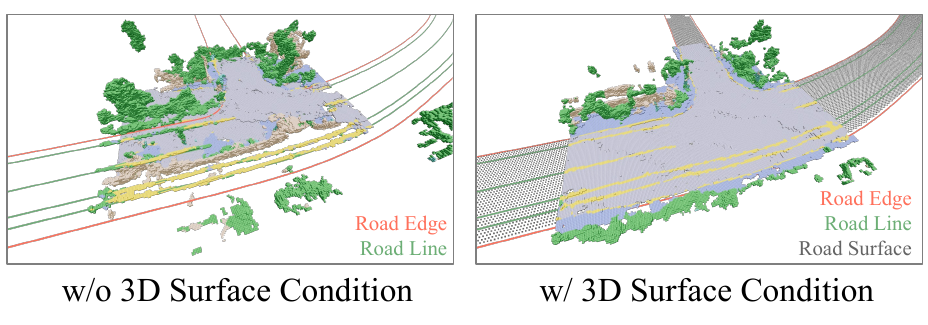}
    \vspace{-1em}
    \caption{\textbf{Impact of 3D Road Surface Condition $\Cond_\text{Road}$.} Without $\Cond_\text{Road}$, the model sometimes fails to determine the ground.}
    \vspace{-1.5em}
    \label{fig:hd_map_abl}
\end{figure}

We perform an ablation study for this component to verify the design of our HD map conditions.
Specifically, in \cref{fig:hd_map_abl} we show the generation results without / with the \emph{Road Surface} condition $\Cond_\text{Road}^\vx$.
One can clearly see the benefits of the condition by helping the network better disambiguate and localize the actual drivable regions.

\begin{figure*}[t]
    \vspace{-0.5em}
    \centering
    \includegraphics[width=0.92\linewidth]{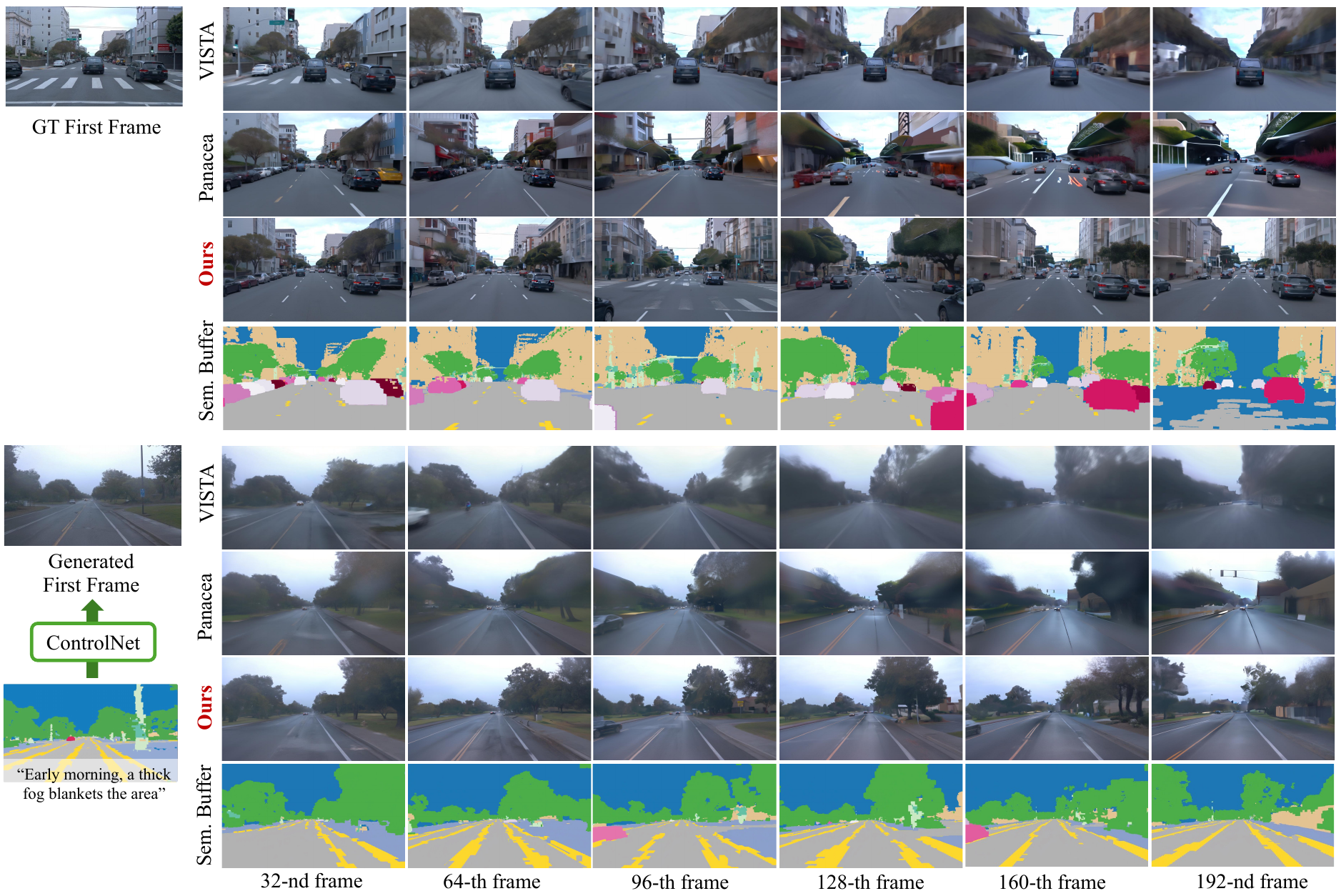}
    \vspace{-0.8em}
    \caption{\textbf{Video Model Comparison.} Our model can generate high-quality videos of \textbf{200 frames} from \textbf{25-frame SVD-XT} by conditioning on guidance buffers. These buffers also enable the motions in the videos to be consistent with the scale of the physical world.}
    \vspace{-1.3em}
    \label{fig:video_model_qualitative}
\end{figure*}

\subsubsection{World-Guided Video Generation}


The proposed 3D grounded guidance buffers significantly maintain high frame quality when generating long videos auto-regressively. To evaluate their effectiveness, we compare with two baselines: Panacea~\cite{wen2024panacea} and Vista~\cite{gao2024vista}, which are both specifically designed for driving video generation.
The original Panacea model is trained based on Stable Diffusion 1.5 --- for a fair comparison, we re-implement their method with the same SVD-XT backbone while adopting their map conditioning strategies, especially their HD map condition that projecting polylines onto the image planes.
Since Vista does not support analytic ego trajectory input, we only use the first frame as its condition without specifying any additional controls.
We utilize the ground-truth first frames from 90 test sequences in the Waymo Open Dataset as conditioning inputs and generate long videos in the same auto-regressive manner. To evaluate the quality of the generated long videos, we compute the Fréchet Inception Distance (FID) at various frame indices.

\begin{figure}[t]
    \centering
    \begin{subfigure}{0.48\linewidth} 
        \centering
        \includegraphics[width=\linewidth]{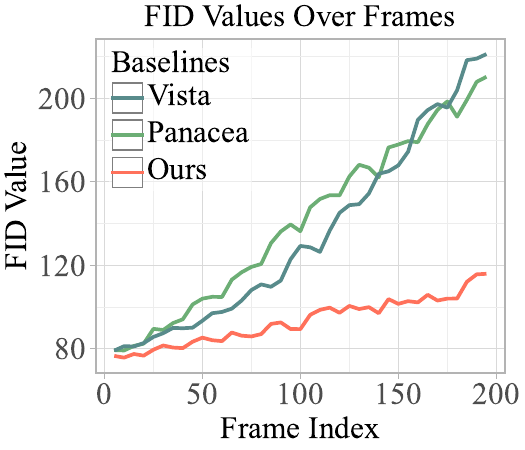} 
        \subcaption{Baseline Comparison on FID.}
        \label{fig:fid_baseline}
    \end{subfigure}\hfill 
    \begin{subfigure}{0.48\linewidth}
        \centering
        \includegraphics[width=\linewidth]{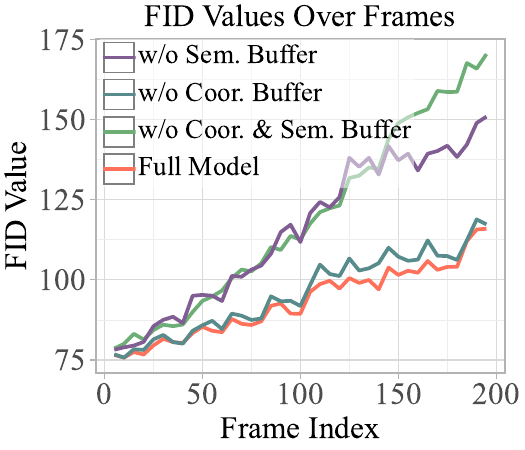} 
        \subcaption{Ablation of Guidance Buffers.}
        \label{fig:fid_ablation}
    \end{subfigure}
     \vspace{-0.7em}
    \caption{\textbf{Comparison of long video generation quality} based on the first frame from the Waymo Dataset. FID: lower is better.}
    \label{fig:combined}
    \vspace{-0.5em}
\end{figure}

Shown in in \cref{fig:video_model_qualitative} and \cref{fig:fid_baseline}, our model maintains a lower FID score and a better visual quality over a long time horizon.
For the baselines, we observe significant quality degradation after $\sim$100 frames, showing the advantage of our 3D grounded buffer conditions.

In \cref{fig:fid_ablation}, we show a detailed ablation study on the effects of different guidance buffers: While the semantic buffer plays the most crucial role in maintaining video quality, the coordinate buffer helps to resolve detailed ambiguities of motion-induced scene changes.

\begin{table}[t]
\setlength{\tabcolsep}{4pt}
\centering
\footnotesize
\begin{tabular}{l|ccc|ccc}
\toprule
 & \multicolumn{3}{c|}{\MethodName (\textbf{Ours})} & \multicolumn{3}{c}{Panacea~\cite{wen2024panacea}} \\ 
\midrule
Frame Index & \texttt{40} & \texttt{80} & \texttt{120} & \texttt{40} & \texttt{80} & \texttt{120} \\ 
\midrule
Positive Rate $\uparrow$ & \cellcolor{first} 84.6\% & \cellcolor{first} 83.9\% & \cellcolor{first} 84.8\% & \cellcolor{second} 76.8\% & \cellcolor{second} 54.0\% & \cellcolor{second} 53.4\% \\ 
\bottomrule
\end{tabular}%
\vspace{-0.8em}
\caption{\textbf{Human evaluation of HD map alignment.} We highlight the \colorbox{first}{best} and the \colorbox{second}{second}.}
\label{tab:user_study}
\vspace{-2em}
\end{table}

To assess the \emph{controllability} of our model, we conducted a user study to evaluate the alignment of our generation w.r.t. the HD map condition. 
We extract the $40^\text{th}$, $80^\text{th}$, and $120^\text{th}$ frames from the video generated by our method and Panacea (sharing the same first-frame generated by ControlNet), and put the map projection beside to ask users if the image aligns with the map. 
A total of 180 image samples are extracted for each frame index, and we report the positive response rate comparison in \cref{tab:user_study}. Results show that our method is more preferential than Panacea at different frames, especially for larger frame indices. 

To summarize, the experiments in \cref{fig:video_model_qualitative}, \cref{fig:fid_baseline}, and \cref{tab:user_study} --- particularly the comparison to Panacea~\cite{wen2024panacea}, which uses HD map projections for conditioning --- demonstrate the advantages of our guidance buffer designs. They not only remarkably reduce auto-regressive errors in long video generation, but also improve HD map alignment in the generated content, which is critical for generating driving training data.

\begin{table}[t]
\setlength{\tabcolsep}{2.2pt}
\begin{center}
\footnotesize
\begin{tabular}{lcccccc}
\toprule
 & \multicolumn{3}{c}{\footnotesize Novel View ($T + 5$)} & \multicolumn{3}{c}{\footnotesize Novel View ($T + 10$)} \\
\cmidrule(lr){2-4} 
\cmidrule(lr){5-7} 
 & PSNR$\uparrow$ & SSIM$\uparrow$ & LPIPS$\downarrow$ & PSNR$\uparrow$ & SSIM$\uparrow$ & LPIPS$\downarrow$ \\
\midrule
PixelNeRF~\cite{yu2021pixelnerf}  & 15.21 & 0.52 & 0.64 & 14.61 & 0.49 & 0.66 \\
PixelSplat~\cite{pixelsplat}   & \cellcolor{third} {20.11} & 0.70 & 0.60 & \cellcolor{third} 18.77 & 0.66 & 0.62 \\
DUSt3R~\cite{wang2023dust3r}  & 17.08 & 0.62 & 0.56 & 16.08 & 0.58 & 0.60 \\
MVSplat~\cite{chen2025mvsplat}   & \cellcolor{second} {20.14} & \cellcolor{third} 0.71 & \cellcolor{third} 0.48 & \cellcolor{second} 18.78 & \cellcolor{third} 0.69 & \cellcolor{third} 0.52 \\
MVSGaussian~\cite{liu2024mvsgaussian}  & 16.49 & 0.70 & 0.60 & 16.42 & 0.60 & 0.59 \\ 
SCube~\cite{ren2024scube}   & 19.90  & \cellcolor{second} {0.72} & \cellcolor{second} {0.47}  & \cellcolor{second} {18.78} & \cellcolor{second} {0.70}  & \cellcolor{second} {0.49} \\
\midrule

\MethodName (\textbf{Ours}) & \cellcolor{first} {20.80} & \cellcolor{first} {0.73} & \cellcolor{first} {0.42} & \cellcolor{first} {19.93} & \cellcolor{first} {0.72} &  \cellcolor{first} {0.45} \\
\bottomrule
\end{tabular}
\end{center}
\vspace{-1.8em}
\caption{\textbf{Quantitative comparisons of novel view rendering.} Metrics are computed at frames $T+5$ and $T+10$ given frame $T$ as input. We highlight the \colorbox{first}{best}, \colorbox{second}{second best} and \colorbox{third}{third best}.}
\label{table:recon_comparison}
\vspace{-0.4em}
\end{table}

\begin{figure}[t]
    \centering
    \includegraphics[width=\linewidth]{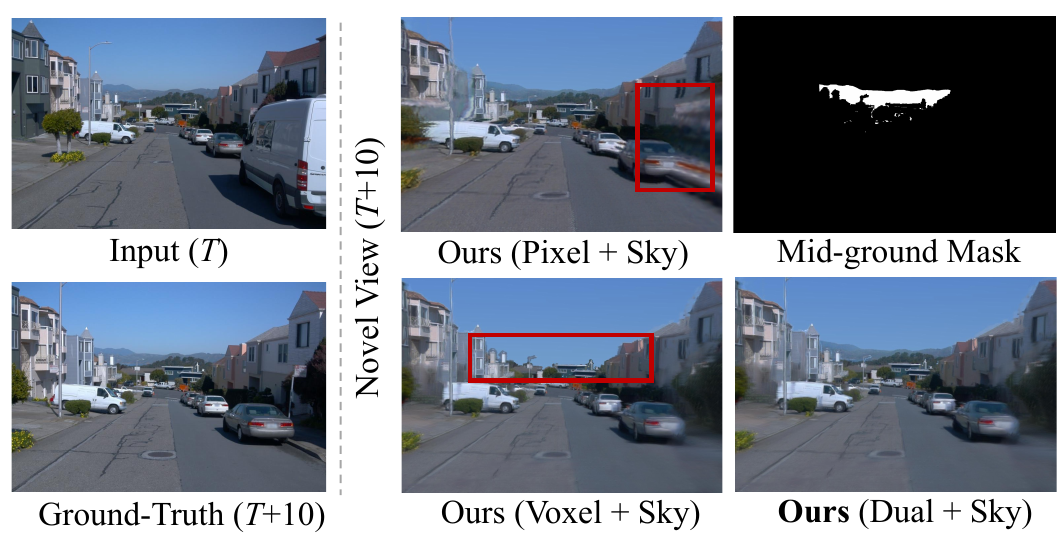}
    \vspace{-0.5em}
    \caption{\textbf{Novel view rendering} of different branches given input images and voxels from Waymo Dataset. Dual inference eliminates the artifacts in either single branch, as shown in the \textcolor{myred}{\textbf{red box}}.}
    \label{fig:gsm_abl}
    \vspace{-1.5em}
\end{figure}

\vspace{-1mm}
\subsubsection{3DGS Scene Generation}
Our 3DGS reconstruction stage works in synergy with the previous two stages to generate high-quality dynamic 3DGS scenes.
To showcase the advantage of our dual-branch reconstruction, we follow the setting from SCube~\cite{ren2024scube} to evaluate the reconstruction quality by synthesizing novel views at frame $T+5$ and $T+10$ given the input views from frame $T$ with 3 front views. 
We show the Peak Signal-to-Noise Ratio (PSNR), Structural Similarity Index Measure (SSIM), and Learned Perceptual Image Patch Similarity (LPIPS)~\cite{zhang2018perceptual} results in \cref{table:recon_comparison}.
Our method outperforms the baselines in all metrics; by introducing the pixel branch, we gain improvement over SCube ~\cite{ren2024scube}.
We further show in \cref{fig:gsm_abl} a qualitative analysis of the renderings coming from different branches of our model.
While each branch has its own artifacts, our dual-branch inference can effectively eliminate them and generate high-quality novel views.

\subsection{Applications}

\begin{figure}
    \centering
    \includegraphics[width=\linewidth]{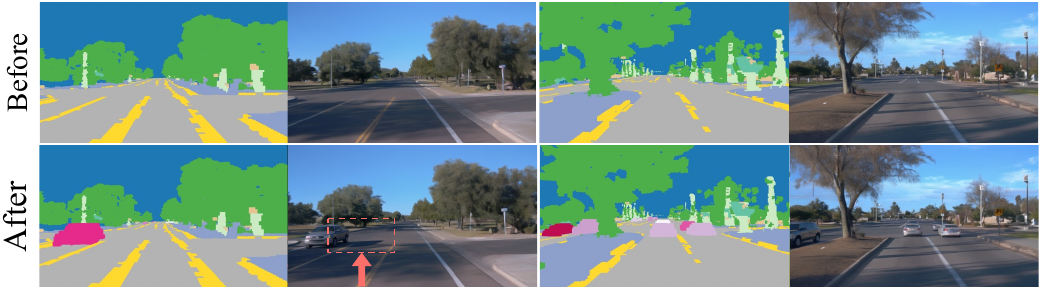}
    \vspace{-1.5em}
    \caption{\textbf{Object insertion by the video model.} We observe realistic shadows cast by the inserted objects (indicated by \textcolor{myorange}{arrow}). }
    \label{fig:object_insertion}
    \vspace{-0.5em}
\end{figure}

With a 3DGS scene, our method naturally supports applications such as novel view synthesis or collision simulation.
Meanwhile, the coherent design of our pipeline also enables more advanced applications as follows:

\parahead{Vehicle Insertion}
New vehicles can be inserted into the scene by simply placing a voxelized car model (with given trajectory) into the voxel world and re-running the subsequent steps.
To maintain the video appearance of the scene before and after the insertion, we keep the first-frame condition unchanged and only update the guidance buffer. We show two object insertion examples in \cref{fig:object_insertion}. 

\begin{figure}
    \centering
    \includegraphics[width=\linewidth]{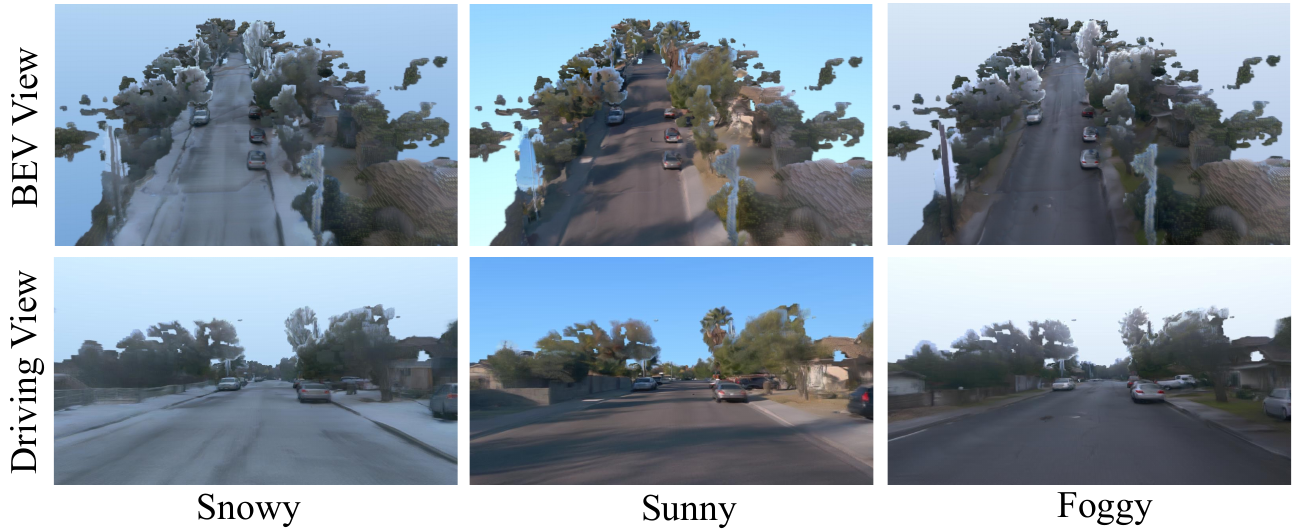}
    \vspace{-1.5em}
    \caption{\textbf{Weather control for scene generation.} We show 3 scenes generated with different text prompts (listed at the bottom).}
    \vspace{-1.2em}
    \label{fig:weather_control}
\end{figure}

\parahead{Weather Control}
Our method allows users to generate scenes with different weather conditions by just altering the text prompts. 
We show 3 scenes with different weather conditions sharing the same underlying voxel world in \cref{fig:weather_control}, where the 3D Gaussians have different appearances.

\vspace{-0.5em}
\section{Discussion}

\vspace{-0.5em}
\label{sec:conclusion}

\parahead{Limitations}%
While the voxel consistency between adjacent chunks is guaranteed by our outpainting strategy, the consistency among long-distance chunks may decline.
Future works include improving global consistency for all chunks and scaling up the model with more diverse training data. 

\parahead{Conclusion}%
In this work, we present \MethodName, a novel method for generating large-scale and high-quality dynamic 3D driving scenes.
Our method is deeply rooted in the synergy among our 3D voxel generation model, the world-guided video model, and the dynamic 3DGS generation model.
Together we can generate realistic 3D scenes with rich appearance details and full controllability.

{
    \small
    \bibliographystyle{ieeenat_fullname}
    \bibliography{main}
}

\clearpage
\onecolumn
\setcounter{section}{0}

\begin{center}
    \vspace*{0.5em}
    \textbf{\LARGE Supplementary Material}
    \vspace*{0.5em}
\end{center}


\renewcommand{\thesection}{\Alph{section}}
\renewcommand{\thetable}{S\arabic{table}}
\renewcommand{\thefigure}{S\arabic{figure}}
\renewcommand{\theequation}{S.\arabic{equation}}

In this supplementary material, we first present further implementation details corresponding to the three main components of our pipeline, \ie, the voxel world generation stage (in \cref{sec:app:voxel}), the world-guided video generation stage (in \cref{sec:app:video}), and the dynamic 3DGS scene generation stage (in \cref{sec:app:recon}).
Additional details about the large-scale generation and user study can be found in \cref{sec:app:inference}, \cref{sec:app:full}, and \cref{sec:app:user_study}. Please also visit the website in the supplementary materials for visualization by clicking the \texttt{index.html}.

\section{Additional Details of the Voxel World Generation}
\label{sec:app:voxel}

\subsection{Voxel Diffusion Model Training}
Following XCube~\cite{ren2024xcube}, we first train a sparse structure Variational Autoencoder (VAE) to encode the semantic sparse voxel grid $\mathbf{D}^{\text{vx}}$ into a dense latent feature cube $\mathbf{X}^{\text{vx}}$, and then train an HD map and 3D bounding box conditioned diffusion model on the latent representation $\mathbf{X}^{\text{vx}}$. 
Here, we do not apply the hierarchical generation since one diffusion is enough for a voxel size of $0.2 \text{m}$ in a range of $51.2\text{m} \times 51.2\text{m}$. 
The diffusion loss is defined with a $\bm{v}$-parameterization: 
{
\begin{equation}
  \mathcal{L}_\text{Diffusion} = \mathbb{E}_{t,\mathbf{X}^{vx},\bm{\epsilon} \sim \mathcal{N}(0, \Identity)} 
  \Bigg[ \left\| \bm{v} \left( \sqrt{\bar{\alpha}_t} \mathbf{X}^{vx} + \sqrt{1 - \bar{\alpha}_t} \bm{\epsilon}, t \right) \right.
  \left. - \left( \sqrt{\bar{\alpha}_t} \bm{\epsilon} - \sqrt{1 - \bar{\alpha}_t} \mathbf{X}^{vx} \right) \right\|_2^2 \Bigg],
\end{equation}
}
where $\bm{v}(\cdot)$ is the diffusion network, $t$ is the randomly sampled diffusion timestamp from $[0, 1000]$, and $\bar{\alpha_t}$ is the
scheduling factor for the diffusion process. More details can be found in ~\cite{ren2024xcube}.

\subsection{Voxel Diffusion Model Sampling}
We use DDIM~\cite{song2020denoising} as our sampler for the distribution of the latent feature cube $\mathbf{X}^{\text{vx}}$ given HD maps and 3D bounding boxes as conditions. We set the denoising step to 100 and the classifier-free guidance~\cite{CFG} weight to 2.0 during inference. To avoid inconsistency from VAE decoding, we do not decode the latent feature cube $\mathbf{X}^{\text{vx}}$ until all the chunks are generated and fused during our outpainting procedure. Then, we use the decoder from the sparse structure VAE to recover the 3D sparse voxel world with the semantic logit for each voxel.

\subsection{Ablation of Outpainting}
\begin{figure}
    \centering
    \includegraphics[width=\linewidth]{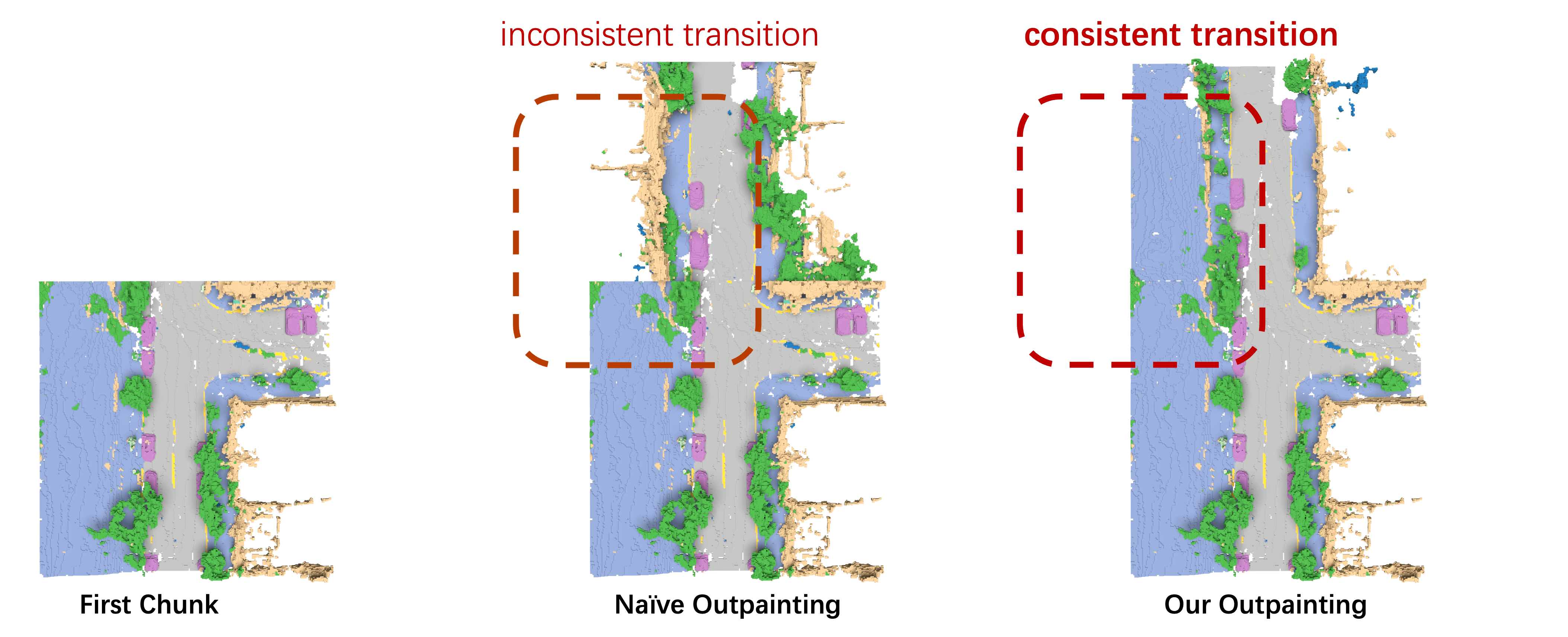}
    \caption{\textbf{Ablation Study of Outpainting Strategy.} Without reusing overlapping latent, there might be inconsistent transitions.}
    \label{fig:outpaint_abl}
\end{figure}

For the voxel outpainting strategy, we compare it with naive outpainting without reusing overlapping latents, showing our outpainting strategy could significantly improve consistency at the transition. See ~\cref{fig:outpaint_abl}.

\section{Additional Details of the World-Guided Video Generation}
\label{sec:app:video}

\subsection{Semantic Buffer Construction}
We show the RGB value of each semantic category in the semantic buffer in \cref{tab:object_categories}. We cluster similar semantic categories together for the coloring.
\begin{table}[ht]
    \centering
        \begin{tabular}{lc}
            \toprule
            \textbf{Semantic Categories} & \textbf{RGB Values} \\ \midrule
            SIGN, TRAFFIC\_LIGHT, CONSTRUCTION\_CONE & \cellcolor[rgb]{0.4, 0.7608, 0.6471} \textcolor{black}{(0.4, 0.7608, 0.6471)} \\
            \begin{tabular}[c]{@{}l@{}}MOTORCYCLIST, BICYCLIST, PEDESTRIAN,\\ \;\;BICYCLE, MOTORCYCLE\end{tabular} & \cellcolor[rgb]{0.9882, 0.5529, 0.3843} \textcolor{black}{(0.9882, 0.5529, 0.3843)} \\
            CAR, TRUCK, BUS, OTHER\_VEHICLE & \cellcolor[rgb]{0.7373, 0.5020, 0.7412} \textcolor{black}{Varying Colors} \\
            CURB, LANE\_MARKER & \cellcolor[rgb]{1.0, 0.8510, 0.1843} \textcolor{black}{(1.0, 0.8510, 0.1843)} \\
            VEGETATION, TREE\_TRUNK & \cellcolor[rgb]{0.3020, 0.6863, 0.2902} \textcolor{black}{(0.3020, 0.6863, 0.2902)} \\
            WALKABLE, SIDEWALK & \cellcolor[rgb]{0.5529, 0.6275, 0.7961} \textcolor{black}{(0.5529, 0.6275, 0.7961)} \\
            BUILDING & \cellcolor[rgb]{0.8980, 0.7686, 0.5804} \textcolor{black}{(0.8980, 0.7686, 0.5804)} \\
            ROAD, OTHER\_GROUND & \cellcolor[rgb]{0.7020, 0.7020, 0.7020} \textcolor{black}{(0.7020, 0.7020, 0.7020)} \\
            UNDEFINED & \cellcolor[rgb]{0.1216, 0.4706, 0.7059} \textcolor{black}{(0.1216, 0.4706, 0.7059)} \\
            POLE & \cellcolor[rgb]{0.8000, 0.9216, 0.7725} \textcolor{black}{(0.8000, 0.9216, 0.7725)} \\ \bottomrule
        \end{tabular}
    \caption{\textbf{Semantic categories and their RGB values in the semantic buffer}. The RGB values in the table range from 0 to 1. In practice, we rescale the above values from -1 to 1 for the encoder.}
    \label{tab:object_categories}
\end{table}

\begin{figure}[h]
    \centering
    \includegraphics[width=0.6\linewidth]{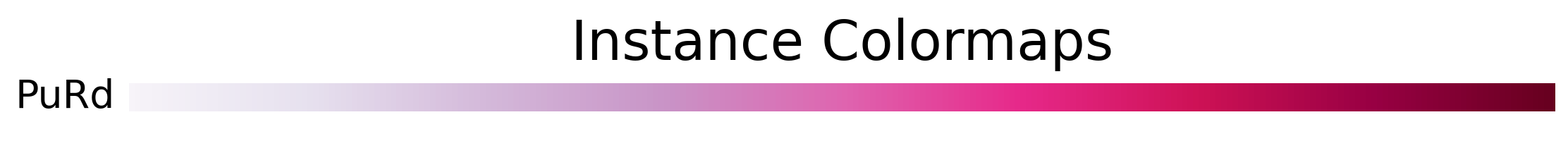}
    \caption{\textbf{Color map for vehicle instances.} We randomly pick a color in \textit{PuRd} color map from Matplotlib for each vehicle instance in the semantic buffer.}
    \label{fig:instance_map}
\end{figure}

Note that we use a varying RGB value across different instances, as mentioned in the main paper for the semantic categories of \texttt{CAR, TRUCK, BUS, OTHER\_VEHICLE}. We randomly select a color in the \textit{PuRd} color map from Matplotlib~\cite{Hunter:2007}, as shown in \cref{fig:instance_map}.

\subsection{First Frame Initialization with ControlNet}

\begin{figure*}[!t]
    \centering
    \includegraphics[width=\linewidth]{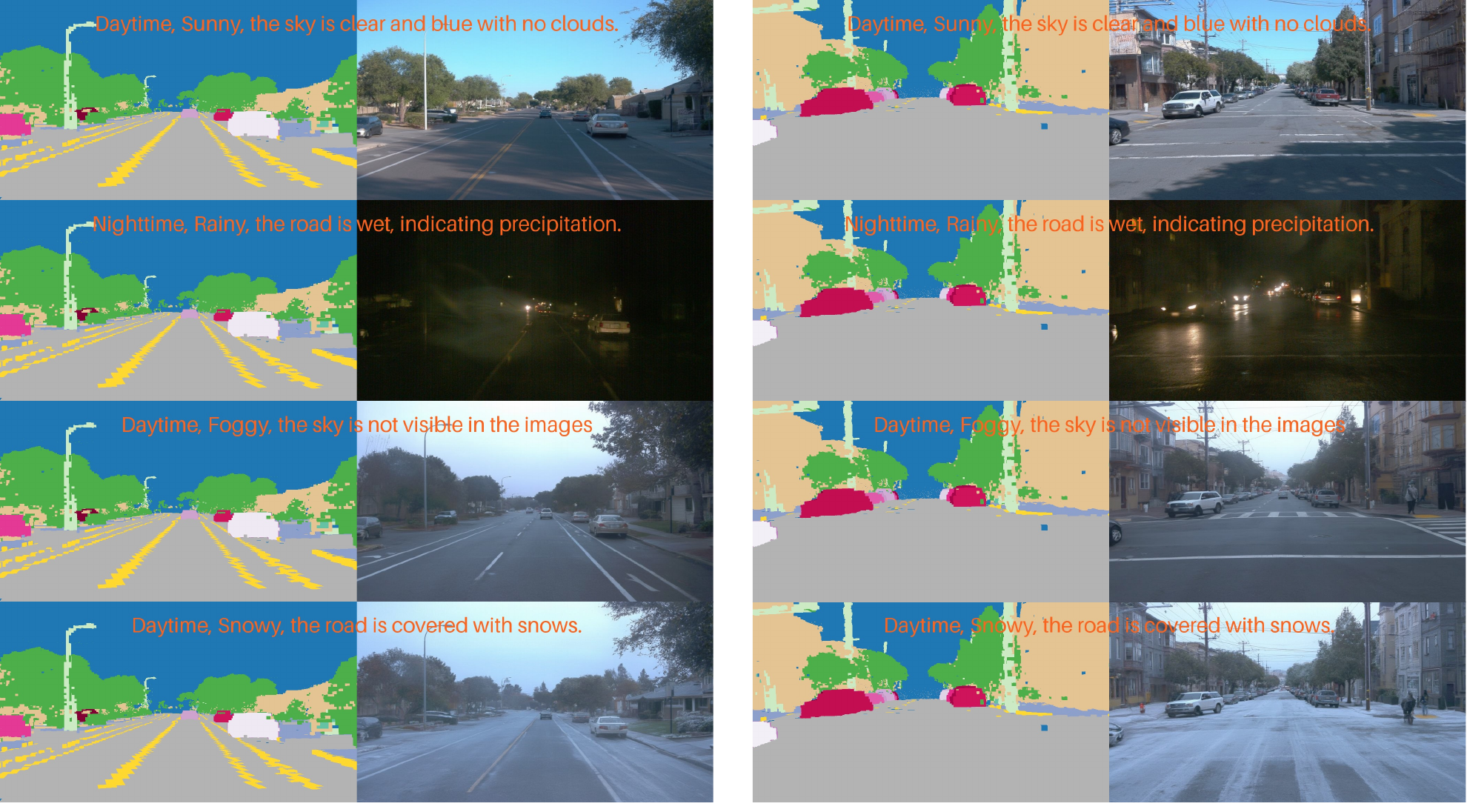}
    \caption{\textbf{First Frame Generation with Different Text Prompts.} We show additional results on generating the initial frame using semantic buffers with ControlNet~\cite{zhang2023adding} based on FLUX~\cite{flux}.}
    
    \label{fig:controlnet}
\end{figure*}

We implement the semantic buffer conditioned FLUX~\cite{flux} ControlNet with the \texttt{diffusers}~\cite{diffusers} library. 
We train the model with an equivalent batch size of 64 for 48 GPU days using NVIDIA A100 GPUs. We show the results of our semantic buffer conditioned ControlNet in \cref{fig:controlnet}.

\section{Additional Details of Dynamic 3D Gaussian Scene Generation}
\label{sec:app:recon}

\subsection{Voxel Branch}
In this work, we adopt a voxel size of $0.2 \text{m}$ in the voxel world generation stage, which is coarser than the voxel size of  $0.1 \text{m}$ used in SCube~\cite{ren2024scube} but is sufficient for video model conditioning in our application.
To further enhance the detail in the scene generation stage, we subdivide the generated voxels into voxel size of $0.1\text{m}$ before the per-voxel Gaussian attribute decoding in the voxel branch. To encode image features, we use several convolutional layers to transform the RGB images $\mathbf{I}$ into 2 $\times$ downsampled feature maps with a channel of 64. We then unproject the feature maps to the voxel world to assign each voxel a voxel feature by max-pooling, and use a 3D UNet to process the 3D voxel feature grid. The final output channel of the 3D UNet model is $56$ for $4$ Gaussians per voxel, where each 3D Gaussian uses 14 channels for RGB $(3)$, rotation $(4)$, scale $(3)$, opacity $(1)$ and relative position $(3)$. The absolute 3D center of the 3D Gaussian is converted from the relative position using the same convention in SCube~\cite{ren2024scube}. 

\parahead{Network Architecture} The 3D UNet~\cite{ren2024scube} has a base channel 64 and a channel multiplier of $[1,2,4]$ for each resolution stage. Within each resolution stage, we use 3D convolutional blocks with kernel size 3 for the feature encoding. 

\subsection{Pixel Branch}
We take original RGB images $\mathbf{I} \in \mathbb{R}^{h\times w\times3}$, randomly masked voxel depths $\tilde{\mathbf{Z}} \in \mathbb{R}^{h\times w\times1}$ and intermediate features $\mathbf{F}_{\text{DAV2}} \in \mathbb{R}^{h\times w\times 32}$ from Depth Anything V2~\cite{yang2024depth} as the input of our 2D UNet model, and predict pixel-aligned 3D Gaussians in this branch. For the randomly masked voxel depth $\tilde{\mathbf{Z}}$, we zero out each non-overlapping $16\times 16$ image patch with a probability of 0.5 in the training stage. Note that we use the full voxel depth $\mathbf{Z}$ in the inference stage (but they still do not cover the mid-ground region). For the Depth Anything V2 feature, we extract the fused feature from the Depth-Anything-V2-\texttt{Large} before the depth prediction head. We then apply several convolutional layers and upsampling layers to resize this fused feature to the image resolution while reducing the number of feature channels to 32. The total input channel for our 2D UNet is $3+1+32=36$. The final output channel of the UNet model is 24 for 2 Gaussians per pixel, where each 3D Gaussian uses 12 channels for RGB $(3)$, rotation $(4)$, scale $(3)$, opacity $(1)$ and depth $(1)$.

We use a similar parameterization for the 3D Gaussians as GS-LRM~\cite{zhang2025gs}. Details for the parameterization can be found in the Appendix of GS-LRM~\cite{zhang2025gs}. The only difference is that we predict depth instead of distance for each Gaussian. This necessitates an additional step to convert the predicted depth into distance to determine the center of the 3D Gaussians in 3D space, utilizing the camera's origin and the ray direction. For a Gaussian $i$, the 3D position is obtained as:
\begin{equation}
\begin{aligned}
    \omega^{i} &= \sigma(\mathbf{G}^{i}_{\text{depth}}), \\
    z^{i} &= (1 - \omega^{i}) \cdot z_{\text{near}} + \omega^{i} \cdot  z_{\text{far}}, \\
    t^{i} &= z^{i} / \cos(\text{ray}_d^{i}, \text{ray}_d^{\text{look-at}}), \\
    \text{xyz}^{i} &=  \text{ray}_o^{i} + t^{i} * \text{ray}_d^{i},
\end{aligned}  
\end{equation}
where the $\mathbf{G}^{i}_{\text{depth}}$ is the model's raw depth output for Gaussian $i$'s prediction, and $\sigma$ is the sigmoid function normalizing the raw output to a weight scalar $w^i$. Here $z^{i}$ is the depth and $t^{i}$ is the corresponding distance;  we set $z_\text{near} = 0.5$ and $z_\text{far} = 300$ in our cases.

\parahead{Network Architecture} The 2D UNet~\cite{unet} has a base channel 32 and a channel multiplier of $[1,2,4,8]$ for each resolution stage. Within each resolution stage, we use 2 ResNet blocks with kernel size 3 for the feature encoding, and an extra Conv/Transposed Conv for the downsampling and upsampling. 

\subsection{Sky Modeling}
We use a lightweight transformer encoder to compress the sky features into a latent feature vector $\mathbf{c} \in \mathbb{R}^{192}$. We prepare a learnable query token $\mathbf{c}_{query}$, similar to the \texttt{[CLS]} token in ViT~\cite{dosovitskiy2020image} for high-level sky feature learning, to interact with all the patches belonging to the sky area (we patchify the image with an $8 \times 8$ patch size and only keep those patches in the sky region). The appearance of the sky will be encoded in $\mathbf{c}$ as follows:
\begin{align}
    \mathbf{c}, \tilde{\mathbf{p}}^{i}_{i\in\{1,2,...,m\}} = \text{TransformerEncoder}(\mathbf{c}_{query}, {\mathbf{p}}^{i}_{i\in\{1,2,...,m\}}),
\end{align}
where $\mathbf{p}^{i}$ and $\tilde{\mathbf{p}}^{i}$ are the sky patches before and after the transformer encoder, $m$ is the number of sky patches. A learning-based positional embedding is applied to the camera ray direction of each patch.

Then we use AdaLN~\cite{karras2019style} to modulate an MLP to take a viewing direction $\mathbf{d}$ and output the corresponding RGB value given the sky vector $\mathbf{c}$. Specifically, we first use learnable embedding to transform the view direction vector to a high-frequency representation $\gamma({\mathbf{d}}) \in \mathbb{R}^{192}$, then normalize the high-frequency view vector $\gamma({\mathbf{d}})$ by LayerNorm without the affine term (we denote the normalized one $\mathbf{x}$). To condition on the sky vector $\mathbf{c}$, we use a linear layer to predict the $\mathbf{scale}$ and $\mathbf{shift}$ from $\mathbf{c}$ for the modulation, i.e.,  $\mathbf{x} = \mathbf{x} \cdot(1 + \mathbf{scale}) + \mathbf{shift}$, and finally decode into a 3-dimensional RGB color with another linear layer. 

\subsection{Additional Application: LiDAR simulation}
\begin{figure}
    \centering
    \includegraphics[width=\linewidth]{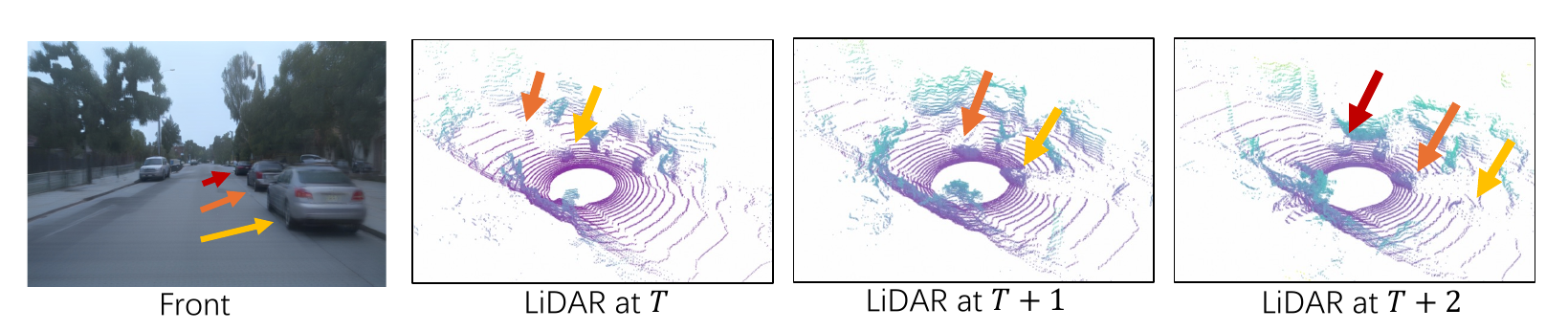}
    \caption{\textbf{LiDAR simulation on generated 3D Gaussian scene.} Ego vehicle is moving forward from timestamp $T$ to $T+2$.}
    \label{fig:lidar_sim}
\end{figure}

With generated 3D Gaussian scenes, we can easily simulate LiDAR points by emitting rays to the Gaussian primitives. See ~\cref{fig:lidar_sim} for illustration.

\section{Inference Speed}
\label{sec:app:inference}
For Stage 1, it takes \SI{30}{\second} and 20GB VRAM to generate a $50\rm m \times 50 \rm m$ local chunk on a single NVIDIA A100 GPU, and about \SI{6}{\minute} to extrapolate to a $30,000 \rm m^2$ voxel world. Stage 2 further takes \SI{8}{\minute} and 75GB VRAM to generate 200-frame video and Stage 3 takes \SI{10}{\second} and 45GB VRAM for reconstruction (both benchmarked on a single NVIDIA A100 GPU). Comparably, VISTA takes more than \SI{12}{\minute} and 70GB VRAM for video generation; and our feed-forward reconstruction runs over 100$\times$ faster than vanilla 3DGS optimization, showing InfiniCube's scalability and efficiency.

\section{Additional Details of Large-scale Scene Generation}
\label{sec:app:full}

We can reuse the ego trajectory from the Waymo Open Dataset~\cite{sun2020scalability} to build the voxel world and generate the guidance buffers. For some parts of the scenes without any coverage of the ego-car trajectories, we use a customized strategy to generate the trajectories for the guidance buffers. While this can be realized by utilizing existing planning modules given HD maps as input, we further implement a real-time viewer based on Viser~\cite{viser2024} for the user to drive in the voxel world. It will record the trajectory and render the voxel scene into guidance buffers. 

\section{Additional Details of User Study}
\label{sec:app:user_study}
\begin{figure}
    \centering
    \includegraphics[width=0.8\linewidth]{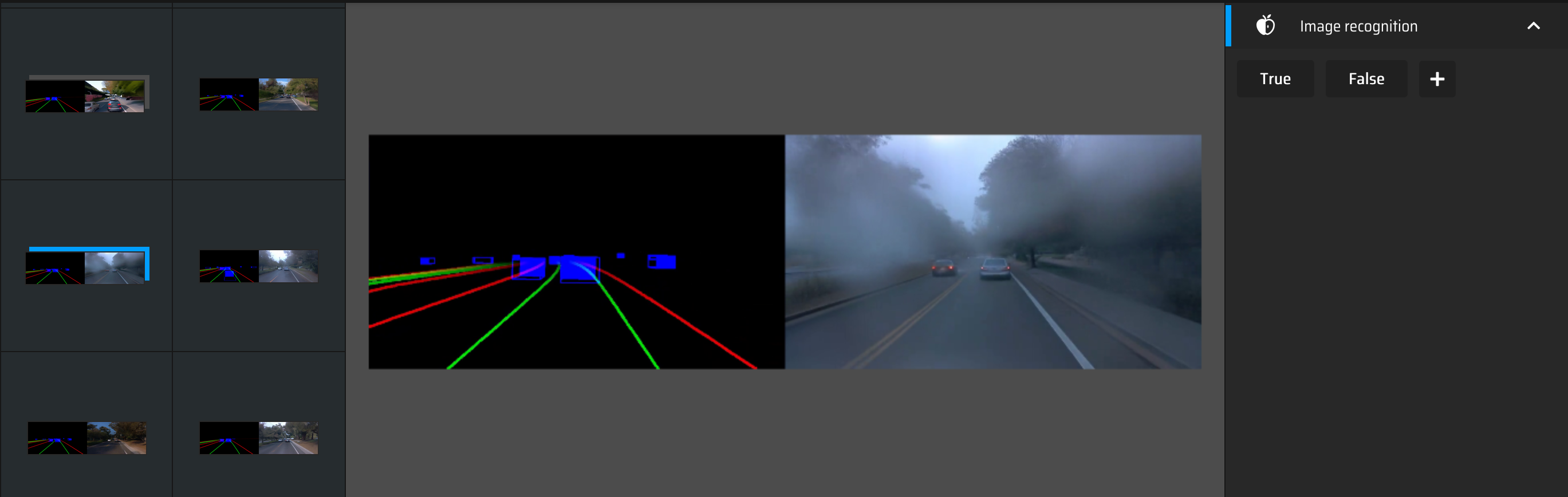}
    \caption{\textbf{User Study Interface Provided by ~\href{https://www.makesense.ai/}{MakeSense AI}}. Users are required to judge if the HD map projection (left) aligns with the RGB image (right). Users are told that the red line is the road boundary, the green line is the lane line, and the blue bounding box is the vehicle.}
    \label{fig:user_study}
\end{figure}

We project the HD map and 3D bounding boxes onto the image plane as a reference for the user to judge if the generated video (image) aligns with the HD map condition.
We generated 63 videos for ours and the baseline methods with different HD map layouts and text prompts, and extracted the $40^{\text{th}}$, $80^{\text{th}}$, and $120^{\text{th}}$ frames from the generated video and ask users to evaluate their alignment. We collected 180 samples for each frame index and calculated their positive rate. The user study is conducted with the platform \href{https://www.makesense.ai/}{MakeSense AI}; see the user interface in Fig.~\ref{fig:user_study}.

\end{document}